# Improving Multi-Word Entity Recognition for Biomedical Texts

**Hamada A. Nayel[1,2], H. L. Shashirekha[2], Hiroyuki Shindo[3], Yuji Matsumoto[3]**

1- Department of Computer Science, Benha University, Benha - 13518, Egypt.

2- Department of Computer Science, Mangalore University, Mangalore - 574199, India.

3- Nara Institute of Science and Technology, Nara - 6300192, Japan.
hamada.ali@fci.bu.edu.eg, hlsrekha@gmail.com, {shindo, matsu}@is.naist.jp

**Abstract**

Biomedical Named Entity Recognition (BioNER) is a crucial step for analyzing Biomedical texts, which aims at extracting biomedical named entities from a given text. Different supervised machine learning algorithms have been applied for BioNER by various researchers. The main requirement of these approaches is an annotated dataset used for learning the parameters of machine learning algorithms. Segment Representation (SR) models comprise of different tag sets used for representing the annotated data, such as IOB2, IOE2 and IOBES. In this paper, we propose an extension of IOBES model to improve the performance of BioNER. The proposed SR model, *FROBES*, improves the representation of multi-word entities. We used Bidirectional Long Short-Term Memory (BiLSTM) network; an instance of Recurrent Neural Networks (RNN), to design a baseline system for BioNER and evaluated the new SR model on two datasets, i2b2/VA 2010 challenge dataset and JNLPBA 2004 shared task dataset. The proposed SR model outperforms other models for multi-word entities with length greater than two. Further, the outputs of different SR models have been combined using majority voting ensemble method which outperforms the baseline model's performance.

## 1. Introduction

Named Entity Recognition (NER) is defined as identifying the named entities (NEs) in the text and classifying them into predefined semantic categories [1]. Names of places, organizations and persons are examples of NEs in general domain, while RNA, DNA, proteins, treatment and medical test are examples of NEs in biomedical domain (BioNEs). Exponential growth of biomedical literature makes it vital to perform BioNER for various applications including Biomedical Text Mining. In addition to general challenges of NER, the nature of BioNEs listed below makes BioNER a challenging task:

1. *Ambiguity*: abbreviations are the major source of ambiguity. A single abbreviation can be interpreted as two different entities according to the context. For example, "EGFR" corresponds to *epidermal growth factor receptor* or *estimated glomerular filtration rate*.
2. *Polysomy*: a word refers to different entities. For example, "*myc-c*" refers to the name of a gene or protein.
3. *Synonyms*: an entity can be denoted by multiple names or aliases. For example, *CASP3*, *caspase-3*, and *CPP32* denote the same entity [2].
4. *Out of dictionary*: the overwhelming growth rate and the frequent insertion of new names into the dictionary [3].
5. *Multi-word BioNEs*: most of BioNEs have multiple words, for example, *CD28 surface receptor*.
6. *Nested BioNEs*: a BioNE may occur as part of longer BioNE as a proper string. For example, "*BP*" *(blood pressure)* corresponding to laboratory test is a BioNE that occurs in "*control BP*" which is a treatment.
7. Lack of standard nomenclature for BioNEs of the same class.

Approaches for BioNER varies from dictionary-based, rule-based, Machine Learning (ML) to hybrid approaches. The widely used ML approaches use annotated data to train a learning model which is then used to classify the unseen BioNEs. Combining the output of different classifiers using ensemble approach is an efficient technique used in BioNER [4]. Ensemble technique tries to overcome the weakness of some classifiers using the strength of other classifiers.

Of late, deep learning algorithms based on Artificial Neural Networks (ANNs) [5-6] are being used to a larger extent to train the learning model for various applications.

## 1.1 Long Short-Term Memory

Artificial Neural Network (ANN) is a programming scheme used to learn the model from observed data. ANN comprises of a large number of interconnected processing units namely, neurons, within different layers. An ANN model basically consists of three layers: input layer, hidden layers and output layer. The input layer contains input neurons that send information to the hidden layer which in turn sends data to the output layer. Recurrent Neural Network (RNN) is a type of ANN in which hidden layer neurons has self-connections which means output depends not only on the present inputs but also on the previous step's neuron state.

Since NER is an instance of the sequence labeling task, it is beneficial to access past and future contexts of sequence tags for NER. RNN architecture is more appropriate to handle sequence data. RNN accepts a sequence of vectors, ($x_1, x_2,...,x_n$) as input and outputs another sequence ($h_1, h_2, ... , h_n$) that contains some information about sequence at every step in the input. In case of long sequences, RNNs are biased towards their most recent inputs in the sequence due to the gradient exploding problem [7-8]. This problem is solved by Long Short-Term Memory (LSTM) [9]. LSTM is a kind of RNN which handles sequences of arbitrary length and is able to model dependencies between far apart sequence elements as well as consecutive elements. Basically, an LSTM unit consists of several gates which control the proportions of information to forget and to pass on to the next step. The complete details of LSTM architecture are described in [9]. One shortcoming of standard LSTM network is that they process the input only in left context, but in NER it is beneficial to have access to both left and right contexts. To overcome this problem, a Bidirectional LSTM (Bi-LSTM) have been designed [10]. The basic idea is to present each sequence forward and backward to two separate hidden states to capture left context and right context information respectively. Then the two hidden states are concatenated to form the final output.

## 2. Segment Representation (SR) Models

One of the major requirements of learning algorithms is an annotated corpus. Segment representation (SR) models which have been applied for different NLP tasks such as Noun Phrase chunking (NP-chunking) [11-12], word segmentation [13-14], NER [15], are more efficient to annotate the data compared to other methods. It is the process of assigning suitable class label(s) to the words in a given text [16]. SR model comprises set of tags, which determine the position of a token in NE, combined with the class label to which that NE belongs to. The tags used in different SR models are **B**, **I, E**, **S** and **O** which stands for **B**egin, **I**nside, **E**nd, **S**ingle and **O**utside respectively. For example, a tag label for a token is B-XXX means that word is the first word of a NE belonging to class XXX. SR model can represent multi-word NEs. Different models are being used to annotate the data.

The primary SR model is **IO** model, it assigns the tag **I** for the tokens inside the entity and the tag **O** for the tokens outside the entity [11]. This model is very simple, but it is not able to represent the boundaries of two consecutive entities of the same class. In the **IOB1** model introduced by Ramshaw and Marcus [17], in addition to tags **I** and **O** the model assigns the tag **B** only to the first token of consecutive NEs of the same class. A modified model of **IOB1** namely **IOB2**, has been introduced by Ratnaparkhi [18]. **IOB2** model assigns the tag **B** for the first word of each NE. The models **IOE1** and **IOE2** use the same concepts of **IOB1** and **IOB2** respectively in addition to using the tag **E** for last token of NE instead of tag **B** [19]. Sun et al. [20], introduced **IOBE** model which concerns with both boundaries of NEs. In addition to traditional tags **I** and **O**, **IOBE** model assigns tags **B** and **E** for the first and last word of all multi-word NEs respectively. **IOBES** model is a modified version of **IOBE** model that concerns with single word NEs. In addition to **IOBE** tags, the **IOBES** model assigns the tag **S** to the single word NEs. This model essentially differentiates between single and multi-word NEs. To show the difference between these models, an example of tagging the text fragment "*The T cell surface molecule CD28 binds to ligands on accessory cells and APCs* ," with different SR models is shown in Table 1.

In this paper, we propose a new SR model to enhance the representation and extraction of multi-word BioNEs. Further, we combine the outputs of different SR models using majority voting ensemble method and the performance of the proposed SR model is evaluated using ANNs architecture based on Bidirectional Long Short-Term Memory (Bi-LSTM) and Conditional Random Fields (CRF).

| Tokens | Segment Representation Models | | | | | | |
|---|---|---|---|---|---|---|---|
| | **IO** | **IOE1** | **IOE2** | **IOB1** | **IOB2** | **IOBE** | **IOBES** |
| The | O | O | O | O | O | O | O |
| T | I-protein | I-protein | I-protein | I-protein | B-protein | B-protein | B-protein |
| Cell | I-protein | I-protein | I-protein | I-protein | I-protein | I-protein | I-protein |
| surface | I-protein | I-protein | I-protein | I-protein | I-protein | I-protein | I-protein |
| molecule | I-protein | E-protein | E-protein | I-protein | I-protein | E-protein | E-protein |
| CD28 | I-protein | I-protein | E-protein | B-protein | B-protein | B-protein | S-protein |
| binds | O | O | O | O | O | O | O |
| To | O | O | O | O | O | O | O |
| ligands | O | O | O | O | O | O | O |
| On | O | O | O | O | O | O | O |
| accessory | I-cell_type | I-cell_type | I-cell_type | I-cell_type | B-cell_type | B-cell_type | B-cell_type |
| cells | I-cell_type | I-cell_type | E-cell_type | I-cell_type | I-cell_type | E-cell_type | E-cell_type |
| And | O | O | O | O | O | O | O |
| APCs | I-cell_type | I-cell_type | I-cell_type | B-cell_type | B-cell_type | B-cell_type | S-cell_type |
| , | O | O | O | O | O | O | O |

**Table 1** Example of Segment Representations

The rest of the paper is organized as follows, Section 3 presents the related works and the details of our proposed model is presented in Section 4. Section 5 contains the details of experiments and results, and the paper concludes in Section 6.

## 3. Related Work

Different ML algorithms such as Support Vector machines (SVMs) [21], Conditional Random Fields (CRFs) [22] and Maximum Entropy (ME) [23] have been used for BioNER. These approaches depend essentially on extracting feature set used for training the appropriate algorithm. Haode et al. [5] have used Convolutional Neural Network (CNN) based model for BioNEs normalization. Xu et al. [6], designed a model using Bidirectional Long Short-Term Memory (Bi-LSTM) and CRF model for clinical named extraction. They used NCBI disease corpus to evaluate their model and have reported a f1-measure of 80.22. Gamal et al. [24] used CNN model for multi-output multi-task model BioNER where they used 15 different datasets for model evaluation, and their model reported an improvement for most datasets.

Lots of research works have been carried out to study the performance of SR models on BioNER [25]. Han-Cheol Cho et al. [26] studied the performance of different SR models using linear chain CRFs to learn a base model for NER. Shashirekha and Nayel [27], studied the performance of BioNER using different SR models. Using CRFs and SVMs for learning the baseline systems for biomedical entity extraction they have compared different SR models on JNLPBA dataset and i2b2/VA 2010 medical challenge dataset. An extension of **IOBES** model has been introduced by Keretna et al. [28] to improve BioNER by introducing a new tag to resolve the problem of ambiguity. The model was evaluated on i2b2/VA 2010 medical challenge dataset.

## 4. Proposed model

### 4.1 FROBES

We propose **FROBES**, an extension of **IOBES** model, used to represent multi-word entities using the tags (F/R/O/B/E/S) for (*front, rear, outside, begin, end, single*) respectively. In this model, the tag **I** in **IOBES** model is replaced by the tags **F** and **R** for entities of length greater than two words. This model describes both halves of the entities, the first half contains tags **B** and **F**, and the second half contains tags **R** and **E**. The relation between the proposed model and other models is shown in Figure 1.

An example of tagging the protein "*human proximal sequence element-binding transcription factor*" using **FROBES** is shown below: -

| *human* | *proximal* | *sequence* | *element-binding* | *transcription* | *factor* |
|---|---|---|---|---|---|
| **B-protein** | **F-protein** | **F-protein** | **R-protein** | **R-protein** | **E-protein** |

***FROBES*** differentiates between the words at the beginning and ending of the multi-word entity. Some multi-word BioNEs have the property of common endings. In many cases, these common ending helps in determining the entity class. For example, many protein names have the common expression "*transcription factor*" at the end of the entity such as;

- *zinc finger* **transcription factor**
- *human proximal sequence element-binding* **transcription factor**
- *B-cell specific* **transcription factor**.

Similarly, many DNA names has the expression "*binding site*" at the end of the DNA name such as;

- *hexameric receptor* **binding site**
- *erythroid Kruppel-like factor (EKLF)* **binding site**

So, expanding tags of multi-word entities to differentiate between both sides of BioNEs may help not only in determining the BioNE, but also to assign it the correct class. The total number of occurrences for tags **F** and **R** in multi-word NEs in **FROBES** assume that the entity consists of *n* words and *n* > 2:

if *n* is even
       # tag **F** is *(n-2)/2*
       # tag **R** is *(n-2)/2*
if *n* is odd
       # tag **F** is *(n-1)/2*
       # tag **R** is *(n-3)/2*

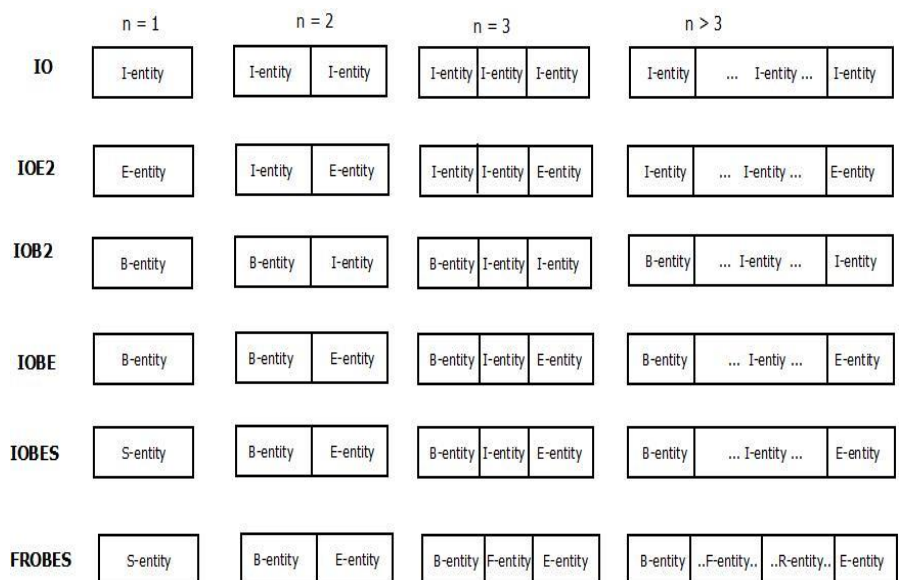

**Fig.1** Relations between Segment Representation models

...

## 4.2 Architecture of baseline model

The structure of the baseline model is shown in Figure 2. Our model accepts a sequence of words and the associated tags as input and gives a contextual representation for each word as output. Each word is represented as two types of vectors namely character embeddings and word embeddings. Character embeddings are used to capture the orthographic features of the words such as capitalization, hyphenation or special characters. Instead of hand-engineering the orthographic information, we learn character embeddings by training a character lookup table initialized randomly with embeddings for every character. The character embeddings corresponding to every character in a word are given in direct and reverse order to BiLSTM, and the output of these BiLSTMs are concatenated to form the character-level representation of a word. This character level representation is then concatenated with word embeddings from lookup-table. Word embeddings vector is used to capture the semantics of words and their similarities based on their surrounding words. We used pretrained word embeddings using skip-gram model induced on a combination of large corpus of PMC[1] and PubMed texts with texts extracted from English Wikipedia dump[2]. This word embeddings model mixes domain-specific texts with domain-independent ones.

At this level, every word is represented as a vector comprising of character level and word level information. Feeding these vectors to a BiLSTM network will output a contextual representation for each word. The final step is decoding, that is converting the contextual representations into output tags. For decoding step CRF is preferable. CRF is an undirected graphical model which focuses on the sentence level instead of individual positions.

## 5. Experiments

We conducted experiments using ANN model which contains a bi-LSTM for character representation and a bi-LSTM for word context representation and CRF for decoding the results to tags.

## 5.1 Performance Evaluation

We used **f1-measure** as a performance evaluation for BioNER system where **TP** is the number of true positives, **FP** number of false positives, and **FN** number of

---

[1] https://www.ncbi.nlm.nih.gov/pmc/
[2] http://bio.nlplab.org/

false negatives and calculated Recall (**R**), Precision (**P**) and **f1-measure** as follows:-

$$P = \frac{TP}{TP + TF}$$
$$R = \frac{TP}{TP + FN}$$
$$f1 - measure = \frac{2 * P * R}{P + R}$$

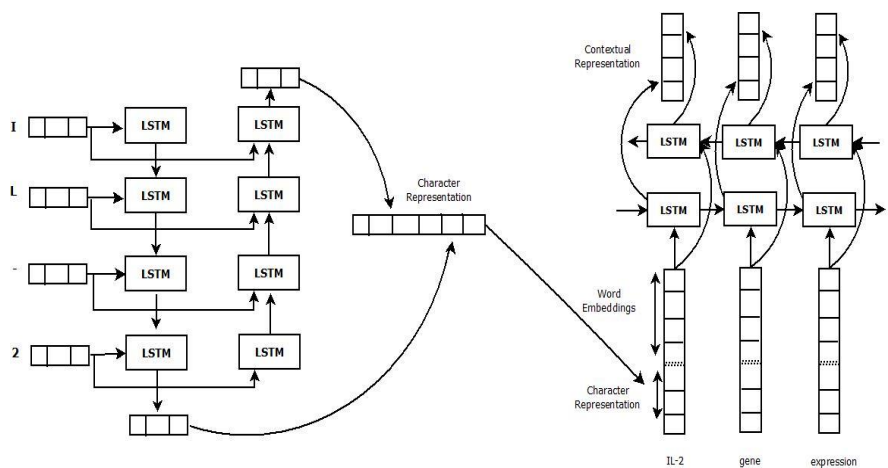

**Fig.2** The structure of baseline system using two BiLSTM neural network

## 5.2. Datasets

In this work, we used two datasets JNLPBA 2004 shared task dataset [29] and i2b2/VA 2010 challenge dataset [30]. Statistic of lengths of named entities of both datasets is given in Table 2.

### 5.2.1 JNLPBA 2004 shared task dataset

The training set is originated from GENIA corpus v3.02 [31]. It consists of 2000 MEDLINE abstracts extracted using the MeSH search terms "*human*", "*blood cell*" and "*transcription factor*". These abstracts were annotated manually into 36 semantic classes. Among these classes, 5 classes are selected in JNLPBA shared task namely *DNA*, *RNA*, *protein*, *cell_line* and *cell_type*. The test set which contains 404 abstracts has been formed using the same MeSH search terms of training set. The publication years for training set ranges over 1990~1999, while for test set it ranges over 1978~2001.

### 5.2.2 i2b2/VA 2010 shared task dataset

This dataset was created for entity and relation extraction purposes at i2b2/VA2010 challenge, including 826 discharge summaries for real patients from the University of Pittsburgh Medical Centre, Partners Health Care and Beth Israel Deaconess Medical Centre. Pittsburgh notes was used as a test set in i2b2/VA 2010 challenge, while other two sources were used as training set. Both test and training sets are manually annotated into three different entities "*treatment*", "*test*" and "*problem*".

| # of words contained in BioNEs | Datasets | | | |
|---|---|---|---|---|
| | JNLPBA | | i2b2 | |
| | Test | Train | Test | Train |
| N = 1 | 3466 40.01 % | 21646 42.19 % | 14116 45.31 % | 7497 45.38 % |
| N = 2 | 2620 30.25 % | 15442 30.10 % | 8469 27.18 % | 4441 26.88 % |
| N = 3 | 1240 14.32 % | 7530 14.68 % | 4573 14.68 % | 2365 14.32 % |
| N > 3 | 1336 15.42 % | 6683 13,03 % | 3996 12.83 % | 2216 13.42 % |
| Total | 8662 | 51301 | 31154 | 16519 |

**Table 2.** Statistics for NEs lengths of JNLPBA and i2b2 datasets

### 5.3 Results and Discussion

The overall results are shown in Table 3. We implemented the base line system with three SR models **IOB2**, **IOBES** and **FROBES**. Also, we used majority voting technique to combine the outputs of these models using ensemble approach. The table shows Recall (**R**), Precision (**P**) and **f1-measure** for implementing baseline system for both datasets using **IOB2, IOBES, FROBES** models and ensemble approach. The results show that our model improves **R** and **f1-measure** for JNLPBA dataset. For i2b2 dataset, **FROBES** model improves **P.** As shown in Table 2 the percentage of entities with length greater than three words in JNLPBA test set is greater than i2b2 test set. **FROBES** model is designed to represent long entities with more appropriate tags. The results show that f-measure of ensemble approach is near to the state-of-the-art for both datasets.

| Datasets | Evaluation Measure | Baseline SR Models | | | Ensemble |
|---|---|---|---|---|---|
| | | IOB2 | IOBES | FROBES | |
| JNLPBA | R | 75.18 | 75.87 | 76.23 | 76.25 |
| | P | 67.82 | 67.68 | 67.69 | 68.16 |
| | f1-measure | 71.31 | 71.54 | 71.71 | **71.99** |
| i2b2 | R | 81.56 | 82.07 | 81.74 | 82.27 |
| | P | 83.84 | 84.57 | 84.62 | 85.01 |
| | f1-measure | 82.68 | 83.30 | 83.15 | **83.62** |

**Table 3.** Results of different SR models and ensemble approach with baseline system

Table 4 and Table 5 illustrate the f-measure of baseline system with different SR models for JNLPBA and i2b2 datasets respectively, with different lengths of entities. It is clear that, **FORBES** outperforms the other two models for multi-word entities (N ≥ 3). Also, ensemble using majority voting improves the f-measure for single and multi-word BioNEs.

| Number of tokens per entity | Baseline SR models | | | Ensemble |
|---|---|---|---|---|
| | IOB2 | IOBES | FROBES | |
| **N = 1** | 73.79 | 73.57 | 73.59 | 73.91 |
| **N = 2** | 74.11 | 74.10 | 74.38 | 74.34 |
| **N ≥ 3** | 64.53 | 65.63 | 65.83 | 66.46 |

**Table 4.** F-measure for JNLPBA dataset

| Number of tokens per entity | Baseline SR model | | | Ensemble |
|---|---|---|---|---|
| | IOB2 | IOBES | FROBES | |
| **N = 1** | 87.66 | 88.04 | 87.92 | 88.23 |
| **N = 2** | 81.64 | 82.21 | 81.63 | 82.31 |
| **N ≥ 3** | 75.58 | 76.82 | 77.00 | 77.48 |

**Table 5**. F-measure for i2b2 dataset

Our model improved the performance of the baseline system for multi-word entity recognition. For single word entities, other models performed better. Our model is designed to discriminate the first part and second part of multi-word entity. In **FORBES**, in addition to tagging inner words as inner tokens in multi-word entities information about position of the word in an entity is also added. We replaced the tag **I** with two tags **R** and **F**. For all tokens at the rear of the entity we use **R**, and **F** for all tokens at the front of the entity. This information helps in improving the learning process.

## 6. Conclusion

We have proposed a new SR model, **FROBES**, to improve multi-word BioNEs representation. To evaluate **FROBES**, we used a Bi-LSTM based model as a baseline system on JNLPBA and i2b2 datasets. Experimental results show that, **FROBES** has improved performance of BioNER for multi-word BioNEs.